\documentclass[conference]{IEEEtran}
\usepackage{graphics}

\usepackage{epsfig}
\usepackage{amsmath}
\usepackage{algorithm}
\usepackage[noend]{algpseudocode}

\ifCLASSINFOpdf
\else
\fi

\begin{document}
%
\title{Stamp processing with examplar features}

\author{Yash Bhalgat \hspace{0.5cm} Mandar Kulkarni \hspace{0.5cm} Shirish Karande \hspace{0.5cm} Sachin Lodha \\ TCS Innovation Labs, Pune, India}

\maketitle

\begin{abstract}

Document digitization is becoming increasingly crucial. In this work, we propose a shape based approach for automatic stamp verification/detection in document images using an unsupervised feature learning. Given a small set of training images, our algorithm learns an appropriate shape representation using an unsupervised clustering. 
Experimental results demonstrate the effectiveness of our framework in challenging scenarios. 
 
%
%
%
%

\end{abstract}


%
\IEEEpeerreviewmaketitle

\section{Introduction}
In developing countries, several transactions take place on paper. In countries like India, there is a strong recent initiative to reduce paper based transaction \cite{ref24}.
Detecting and verifying stamps in documents is an important problem since stamps can be indicators of authenticity. 

In this paper, we propose a shape based stamp verification/detection approach for Indian document stamps. We resort to an unsupervised feature learning approach for learning an appropriate representation for stamp shapes.
Recently, there has been a study that the single layer of convolution filters learned with an unsupervised dictionary learning method such as K-means clustering performs well on object recognition \cite{ref1}. 
The accuracy of object recognition improves with more number of dictionary atoms. 
However, the significance or contribution of each dictionary atom towards the final recognition rate is not reported. We demonstrate that the high recognition rates can be obtained even with less number of dictionary atoms chosen \emph{carefully}. We propose an atom ranking scheme which then automatically selects the dictionary atoms which are indeed useful for good performance.

\begin{figure} [!h]
\centering
\begin{tabular}{c c c c}
\includegraphics[height = 30pt,width = 50pt]{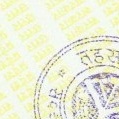}&
\includegraphics[height = 30pt,width = 50pt]{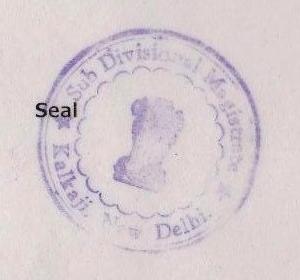}&
\includegraphics[height = 30pt,width = 50pt]{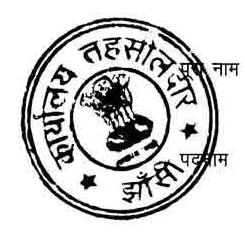}&
\includegraphics[height = 30pt,width = 50pt]{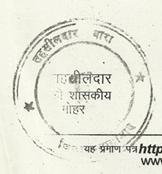}\\
\includegraphics[height = 30pt,width = 50pt]{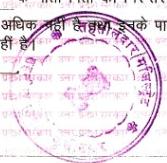}&
\includegraphics[height = 30pt,width = 50pt]{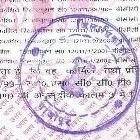}&
\includegraphics[height = 30pt,width = 50pt]{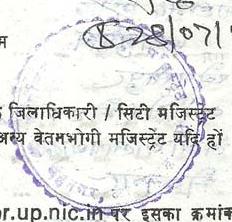}&
\includegraphics[height = 30pt,width = 50pt]{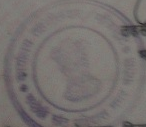}\\
\end{tabular}
\caption{\label{fig:st1} Example images from our scanned document dataset.}
\end{figure}

We performed experiments on our propriety dataset of scanned caste certificate documents. 
Due to no restriction enforced on scanning type, a document may or may not contain color which renders color based approaches not usable.
Fig. \ref{fig:st1} shows example stamp images from our dataset. Our stamp dataset suffers from issues such as faded/poorly imprinted stamps, stamp-text overlap, poor scanning quality, low resolution, time degradations which renders recognition non-trivial. High recognition rates reported in experimental results demonstrate efficacy of our method. Our approach also out-performs off-the-shelf shape descriptors such as Gabor filters.

\section{Our methodology}

\subsection{Training data generation}
Training data for stamp images was obtained through a crowd-sourcing experiment where each worker was asked to draw a box around stamp. 
Due to inter-worker variability, the box markings were non-uniform. Stamp data thus suffers from issues such as partial markings, translation and margin variations as can be seen in Fig. \ref{fig:st1}.



\subsection{Feature learning and extraction}
Feature representation for stamp is learned as following. 

\begin{itemize}

\item Randomly sample patches of size $m \times m$ from stamp images
\item Perform ZCA whitening on patches
\item Perform $K$-means clustering to obtain dictionary atoms
\item Rank dictionary atoms as described in section \ref{rank}

\end{itemize}

Using the learned dictionary atoms, from an image, features are extracted as following.
\begin{itemize}
\item Convolve an image with learned dictionary atoms 
\item Use 1-of-$K$, max-assignment for encoding as follows
\[
    f_K(x)= 
\begin{cases}
    f_K(x), & \text{if } K = \arg \max f(x)\\
    0,              & \text{otherwise}
\end{cases}
\]
\item Perform $4 \times 4$ - quadrant max pooling on the feature maps
\item Form a feature vector by concatenating features
\end{itemize}

\begin{figure} [!h]
\centering
\begin{tabular}{c c}
\includegraphics[height = 100pt,width = 100pt]{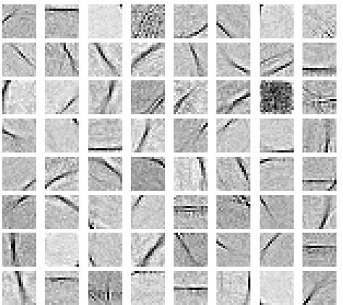}&
\includegraphics[height = 100pt,width = 100pt]{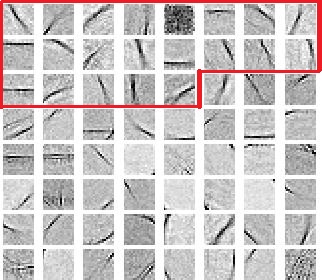}\\(a)&(b)\\
\end{tabular}
\caption{\label{fig:st2} $K$-means clustering result: (a) Learned dictionary, (b) ranked dictionary atoms. Red marking shows the subset of ranked dictionary atoms picked. }
\end{figure}

Fig. \ref{fig:st2}(a) shows the learned dictionary ($D$) where $K = 64$. 

Note that most of the dictionary atoms exhibit the directional nature, however, there are atoms which portrays almost a flat region and are less informative. This can happen because of random sampling of patches where not only stamp regions but also patches from the background get picked.  
To identify the dictionary elements which are most useful for recognition, we propose a dictionary atom ranking scheme.
\subsection{Ranking dictionary atoms}\label{rank} 
We randomly pick a stamp image from our training set. From the training image, overlapping patches of size $m \times m$ are obtained from all pixel locations (i.e. stride is set to 1). Let $Y$ denotes the patch set.
We project $Y$ on the obtained $K$ atoms and perform thresholding using a Rectified Linear unit (ReLu) as follows
\begin{eqnarray}\label{eq:b}
R_{ij} = (1 - y_{ic}) \max (0 , D_j^{T} \hspace{0.1cm} y_i ) \hspace{0.2cm} i \in [1,n]
\end{eqnarray}
where $R_{ij}$ denotes the response of $j^{th}$ atom for $i^{th}$ patch and $n$ denotes the number of patches in $Y$. 
$y_{ic}$ denotes the intensity value at the center of the patch. 
Since stamps are on a lighter background, post multiplication by $(1 - y_{ic})$ assigns more weight to the patch response if it contains a part of stamp. 
The above operation is equivalent to convolving $K$ filters with the training image, performing rectification on the result and pixel-wise multiplying by an inverted input image. 
Response for a dictionary atom is calculated as the maximum of an overall response.
\begin{eqnarray}\label{eq:c}
S_j = \max_i R_{ij}
\end{eqnarray}
where $S_j$ denotes the maximum response attained by $j^{th}$ atom. We rank the atoms in the descending order of their responses. Fig. \ref{fig:st2}(b) shows the ranked atoms. Note that the atoms which partly represent the circular shape are ranked higher than the rest. 
\emph{An interesting observation}: it may appear that the fifth atom in the first row of Fig. \ref{fig:st2}(b) does not show directional nature. We note that it actually represents an emblem which appears at the center of most of the stamps.   
We then chose top $v$ atoms to be used for sub-sequent processing. The value for $v$ is chosen based on a pre-defined threshold on the maximum response. The red boundary in Fig. \ref{fig:st2}(b) shows the atoms which are picked in the process.

\section{Experimental results}
In this section, we demonstrate results of our method for stamp verification and stamp detection.
\subsection{Stamp verification}
Given a test image, our aim is to classify it as a stamp or non-stamp.
For obtaining the dataset for non-stamp images, we use the fact that stamps in our documents always lie in the lower half side. We, therefore, randomly sample patches from the upper half only. Our non-stamp set mainly consisted of text regions, background regions or document borders. 
Our training data thus consist of 882 stamp and 957 non-stamp images. Prior to feature extraction, all the images are converted to grayscale, resized to a fixed dimension and normalized in the range 0 to 1. We use the patch size of $16 \times 16$ for our experiments.   
The feature set is randomly divided in 70\%-30\% for training and testing respectively.
We train a binary linear SVM classifier on training features and compute classification accuracy on the test set. 
For comparison, we performed the classification with following settings: subset of ranked dictionary atoms ($v = 21$), use all dictionary atoms ($v = 64$), 64 Gabor filters (8 scale and 8 orientations), 64 Random Filters (RF). 
Table \ref{Table:res} shows our classification results. Note that, a small set (approx. $\frac{1}{3}$rd) of ranked dictionary atoms produces a slightly superior performance as compared to the full set (with less testing time). Testing time reported here is with MATLAB implementation. We also observe that our approach significantly outperforms off-the-shelf shape descriptor such as Gabor filters and a single layer of random filter based recognition. 

\begin{table}[h]

\sffamily
	\centering
		\begin{tabular}{|p{1.4cm}|p{1.4cm}|p{0.8cm}|p{0.8cm}|p{0.8cm}|p{1cm}|}		
		  \hline
			\textbf{Method} & \textbf{\# of filters}  & \textbf{Acc.} & \textbf{Prec.} & \textbf{Recall} & \textbf{Test time (s)}\\
			\hline
			K-means & 21 & \textbf{94.57} &	100 & 90.57 & 0.88 \\
			\hline
			K-means & 64 & 94.2 &	99.57 & 88.3 & 2.414 \\
			\hline
			Gabor & 64 & 90.22 &	100 & 82.26 & 2.54 \\
			\hline
			RF & 64 & 76.09 &	96.5 & 52.08 & 2.66 \\
			\hline
			\end{tabular}
	
	\caption{\label{Table:res} Experimental results.}
			
\end{table}


%
%
%
%
%
%
%
\vspace{-0.7cm}

\subsection{Stamp detection}
\begin{figure} [!h]
\centering
\begin{tabular}{p{1.7cm} p{1.7cm} p{1.5cm} p{1.5cm}}
\includegraphics[height = 80pt,width = 70pt]{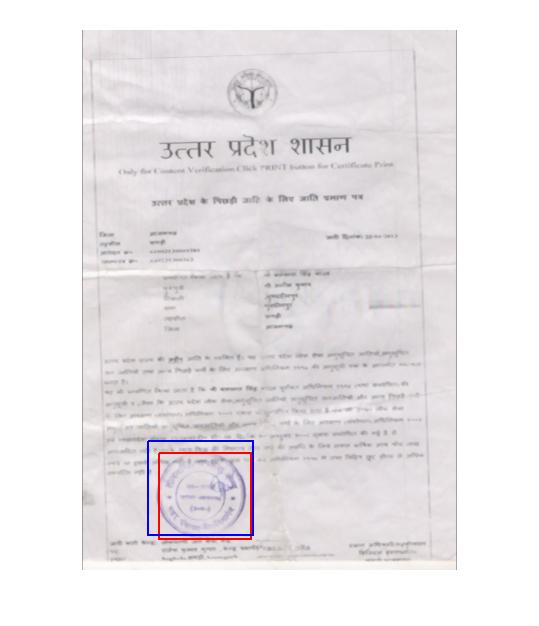}&
\includegraphics[height = 80pt,width = 70pt]{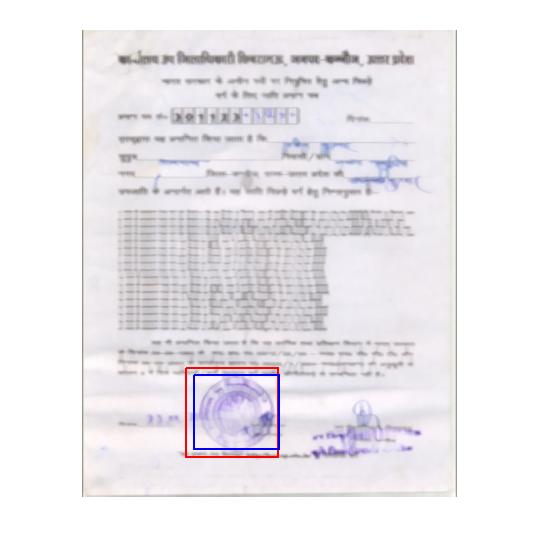}&
\includegraphics[height = 80pt,width = 70pt]{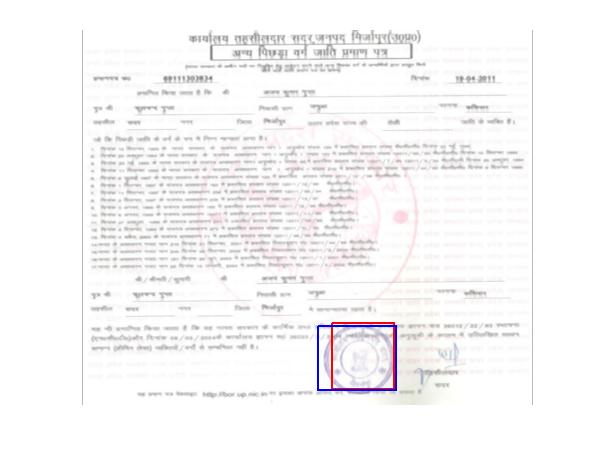}&
\includegraphics[height = 80pt,width = 70pt]{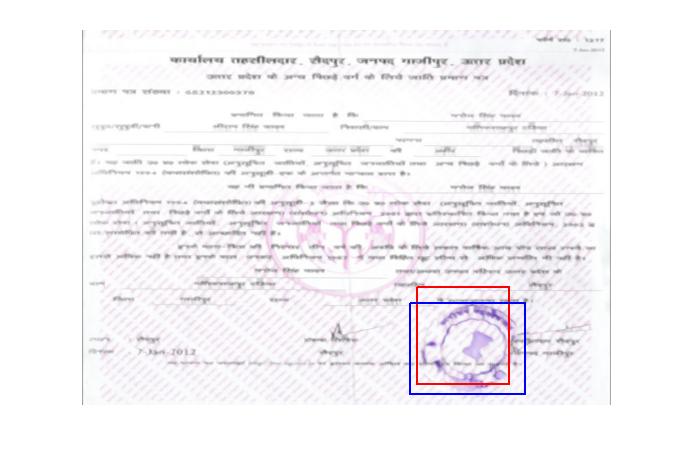}\\
\end{tabular}
\caption{\label{fig:st5} Stamp detection results: Blue box shows the ground truth while red box shows the estimated bounding box.}
\end{figure}

The subset of ranked filters can also be used to locate (segment) stamps from images. 
We convolve the top $v$ filters with the input image and perform rectification as per Eq. \ref{eq:b}.
We compute an average of the responses from the filters. It is observed that, we get a relatively high response at the stamp locations and a low response at non-stamp locations. 
Using a moving window sum method, a region of maximum response is located. Bounding box of the stamp is then decided by local threshold based heuristic method. 
Stamp detection performance is measured as an average Intersection over Union (IoU) overlap between the box markings obtained from the crowd-sourcing experiment and ones which are estimated algorithmically. 
We get an average IoU overlap of 74.81\% which underlines efficiency of our method.
Fig. \ref{fig:st5} shows examples of our detection results.

\section{Conclusion}
In this paper, we proposed an unsupervised feature learning based approach for stamp detection and verification. We have demonstrated that the subset of ranked dictionary atoms provides a better performance with less computing. We also proposed a scheme to rank and choose the subset. Experimental results showed an effectiveness of our method.


\end{document}